\documentclass[sigconf]{acmart}

\usepackage{amsmath,amsfonts}
\usepackage{algorithmic}
\usepackage{algorithm}
\usepackage{array}
\usepackage{subcaption}
\usepackage{textcomp}
\usepackage{stfloats}
\usepackage{url}
\usepackage{verbatim}
\usepackage{graphicx}

\def\etal{\textit{et al.}}

\usepackage{booktabs}
\usepackage{amsmath}
\usepackage{bm}
\usepackage{wrapfig}
\usepackage[dvipsnames]{xcolor}
\usepackage{orcidlink}
\usepackage{pifont}
\usepackage{multirow}

\usepackage{colortbl}
\definecolor{tableHeadGray}{gray}{.9}
\definecolor{tableSubHeadGray}{gray}{0.95}

\newcommand{\Skip}[1]{}

\AtBeginDocument{%
  }

\setcopyright{acmlicensed}
\copyrightyear{2026}
\acmYear{2026}
\setcopyright{cc}
\setcctype{by}
\acmConference[MM '26]{Proceedings of the 34th ACM International Conference on Multimedia}{November 10--14, 2026}{Rio de Janeiro, Brazil}
\acmBooktitle{Proceedings of the 34th ACM International Conference on Multimedia (MM '26), November 10--14, 2026, Rio de Janeiro, Brazil}
\acmDOI{10.1145/3767308.3835301}
\acmISBN{979-8-4007-2213-4/2026/11}

\begin{document}

\title{EvBS: Event-guided Blur Synthesis for Domain-adaptive Motion Deblurring}

\author{Junsik Jung}
\email{junsik.jung@kaist.ac.kr}
\affiliation{%
  \institution{Korea Advanced Institute of Science and Technology}
  \city{Daejeon}
  \country{Republic of Korea}
}

\author{Seokryun Choi}
\email{srchoi99@kaist.ac.kr}
\affiliation{%
  \institution{Korea Advanced Institute of Science and Technology}
  \city{Daejeon}
  \country{Republic of Korea}
}

\author{Yoonki Cho}
\email{yoonki@kaist.ac.kr}
\affiliation{%
  \institution{Korea Advanced Institute of Science and Technology}
  \city{Daejeon}
  \country{Republic of Korea}
}

\author{Woo Jae Kim}
\email{wkim97@kaist.ac.kr}
\affiliation{%
  \institution{Korea Advanced Institute of Science and Technology}
  \city{Daejeon}
  \country{Republic of Korea}
}

\author{Andrew Jeong}
\email{andrew7447@kaist.ac.kr}
\affiliation{%
  \institution{Korea Advanced Institute of Science and Technology}
  \city{Daejeon}
  \country{Republic of Korea}
}

\author{Sung-Eui Yoon}
\authornote{Corresponding author.}
\email{sungeui@kaist.edu}
\affiliation{%
  \institution{Korea Advanced Institute of Science and Technology}
  \city{Daejeon}
  \country{Republic of Korea}
}

\renewcommand{\shortauthors}{Jung et al.}


\begin{abstract}
Motion deblurring has achieved remarkable progress with deep learning, yet pre-trained deblurring models often suffer from performance degradation in real-world scenarios due to the domain shift between training and testing distributions.
To remedy this, we propose EvBS, an event-guided blur synthesis framework that generates diverse training pairs for calibrating pre-trained models to the target domain.
While existing methods are constrained by the inherent entanglement between motion and visual content, our method leverages the high temporal resolution of event cameras to effectively decouple them. 
This enables us to utilize not only the intrinsic motion that is inherent to the given content but also extrinsic motion transferred from different sources within the target domain, thereby facilitating effective adaptation via fine-tuning.
Specifically, EvBS comprises two complementary strategies: Intrinsic-Blur Synthesis, which blurs sharp contents with their own motion patterns, and Extrinsic-Blur Synthesis, which transfers motion from blurry patches to distinct sharp content. 
This approach generates a diverse set of training pairs that break the inherent constraints of naturally coupled motion and content, resulting in enhanced domain-adaptive deblurring performance.
Extensive experiments on multiple benchmarks demonstrate that EvBS effectively enhances the robustness of existing deblurring models on unseen testing datasets.
\end{abstract}




\maketitle

\begin{figure}[!t]
    \centering
    \includegraphics[width=0.48\textwidth]{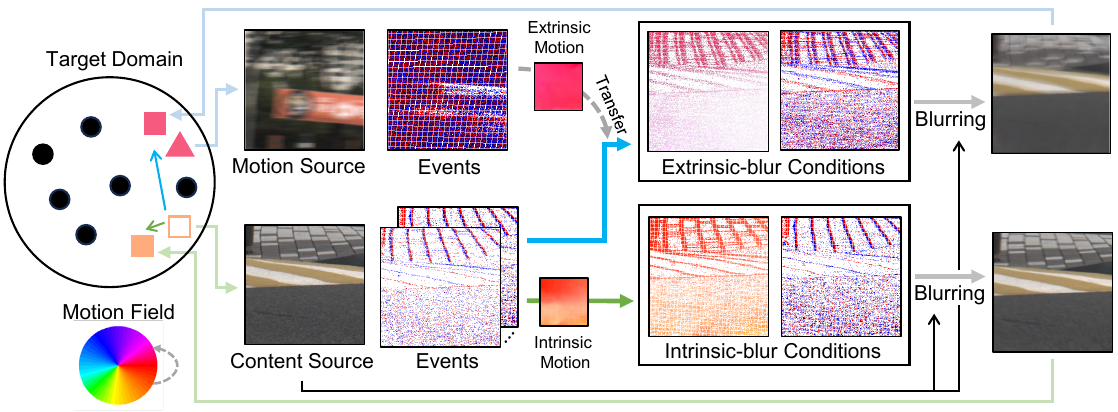}
    \vspace{-4mm}
    \caption{
    Illustration of EvBS.
    In the target domain representation, shapes denote visual content and colors represent distinct motion patterns. By leveraging events to explicitly decouple motion from content, EvBS enables both intrinsic blur synthesis (applying a content source’s inherent motion) and extrinsic blur synthesis (transferring motion from a separate motion source). This strategy generates a diverse set of synthesized training pairs within the target domain, facilitating enhanced model calibration.
    }
    \vspace{-5mm}
    \label{fig:intro}
\end{figure}

\section{Introduction}
\label{sec:intro}

Motion deblurring, the task of restoring a sharp image from a blurry input caused by camera or object movement, is a fundamental challenge in computational photography. While learning-based methods have driven significant progress~\cite{ImageDeblurSurvey, mimounet, MultiScaleDiffusion}, their reliance on standard RGB frames inherently limits their performance in highly dynamic scenes. 
In conventional cameras, motion and visual content are inextricably entangled within a single exposure, causing a significant loss of motion information that makes it difficult to handle severe blur.
To overcome this limitation, recent research has increasingly adopted event cameras~\cite{AsyncVisionSensor}. Thanks to their microsecond-level temporal resolution and high dynamic range, event streams provide highly reliable motion cues, leading to a new paradigm of event-guided deblurring architectures~\cite{efnet, EIFNet, EBFI}.

Despite these advances, a critical challenge persists: the domain shift. Pre-trained deblurring models often exhibit significant performance degradation when applied to real-world test scenes that differ from their training distribution~\cite{realblurdataset}. This discrepancy, typically arising from variations in camera environments and motion blur distributions, poses a significant risk to applications requiring reliable perception, such as autonomous driving and robotics~\cite{BenchmarkingRobustness}. To address this, several studies~\cite{MetaAux, CleanHard} have introduced test-time adaptation approaches that derive supervision signals by reconstructing the input blurry image during inference. However, such indirect supervision often provides insufficient guidance for robust adaptation, restricting the model to limited update steps and yielding only marginal performance gains.

Moving a step further, recent domain-adaptive schemes~\cite{metaTransfer, DADeblur} synthesize training pairs for fine-tuning the pre-trained models by identifying pseudo-sharp patches in the test data and blurring them with their estimated motions. While promising, these methods face the following limitations:
(1) They are constrained by the inherent entanglement between motion and visual content in RGB frames — meaning motion trajectories are baked into captured consecutive frames and are highly challenging to disentangle. Consequently, these methods can only re-apply a motion pattern back to its original pseudo-sharp counterpart. 
This inability to mix-and-match diverse motions with different contents restricts the diversity of the synthesized data, hindering robust adaptation to various real-world blurs.
(2) They are designed exclusively for frame-based architectures, making them incompatible with recent event-based deblurring models~\cite{efnet, maenet, EIFNet, EBFI}. In light of the increasing adoption of events for robust multimedia perception~\cite{event_vision_survey}, establishing adaptation frameworks for event-based architectures is highly necessary.

To overcome these challenges, we propose \textbf{EvBS}, an \textbf{Ev}ent-guided \textbf{B}lur \textbf{S}ynthesis framework that effectively generates diverse blurred patches within the target domain (See Fig.~\ref{fig:intro}). By decoupling motion from content via the high temporal resolution of event cameras, our method not only performs \textit{intrinsic blur} synthesis, which uses the native motion pattern of a content source, and \textit{extrinsic blur} synthesis, which transfers motion patterns from other patches to the given content.
Specifically, EvBS comprises three main components:
(1) A \textbf{Dual Source Extractor} that scans the target domain to locate sharp content sources and motion-consistent blur sources.
(2) Operating directly on these detected sources, a \textbf{Blur Conditioning Module} generates blur conditions via two generators: \textit{Intrinsic-blur Condition Generator} that extracts a sharp content source's inherent motion, and \textit{Extrinsic-blur Condition Generator} that applies motion pattern from the blur sources onto another sharp content.
(3) These conditions guide a diffusion-based model~\cite{IDBlau} to synthesize diverse training pairs for fine-tuning. By utilizing diverse motion patterns inherent in the target domain, we enable a much broader spectrum of training data synthesis. Moreover, our method is broadly applicable to both event-based and frame-based deblurring models.

Our main contributions are summarized as follows:
\begin{itemize}
    \item We propose EvBS, a novel event-guided blur synthesis framework for domain-adaptive motion deblurring. By leveraging events to explicitly decouple motion from visual content, EvBS generates a diverse set of training pairs, breaking the inherent constraints of naturally coupled motion and content.
    
    \item We introduce a framework featuring a Dual Source Extractor and a Blur Conditioning Module to enable two complementary strategies: Intrinsic-Blur Synthesis for applying inherent motion, and Extrinsic-Blur Synthesis for transferring motion across distinct sources.
    
    \item Extensive experiments demonstrate that EvBS effectively enhances the robustness of both pre-trained image- and event-based deblurring models on unseen real-world datasets.
\end{itemize}

\vspace{-3mm}
\section{Related Work}
\subsection{Motion Deblurring}
Early foundational works on motion deblurring have predominantly focused on CNN-based architectures~\cite{ImageDeblurSurvey}. These methods aim to learn a direct mapping from a blurry input to its corresponding sharp image. To enhance performance, the field has seen a variety of innovations, including the integration of attention mechanisms~\cite{SpatialAttentionMD, BANet, stripformer}, coarse-to-fine strategies~\cite{mimounet, WaveletMultiScaleMD}, and diffusion-based approaches~\cite{MultiScaleDiffusion, HierarchicalDiffusion}. These advancements have significantly improved the accuracy and robustness of deblurring models on established benchmarks. However, image-based approaches inherently face limitations when dealing with severe motion blur, as they struggle to leverage the rich temporal information embedded within the blur itself.

To overcome the limitations of image-based methods, recent research has shifted towards using novel sensors, most notably the event camera~\cite{AsyncVisionSensor}, whose high temporal resolution is particularly advantageous for motion deblurring.
These approaches are generally built upon event-based image formation models~\cite{edi} and integrate them with deep learning techniques to achieve remarkable efficiency and accuracy.
Seminal works in this area have introduced sophisticated event representation and attention mechanisms to effectively fuse image and event features~\cite{efnet, maenet, EIFNet}. Furthermore, the field has progressed towards addressing more realistic and challenging constraints, such as deblurring under blind exposure times~\cite{uevd, EBFI}.

Despite these advancements in both image-based and event-based methods, a common challenge persists across the field: many of these methods exhibit a significant performance drop when applied to unseen domains, particularly on real-world blurry images that differ from their training distributions. This domain gap issue motivates the need for domain adaptation techniques in motion deblurring.


\vspace{-3mm}
\subsection{Blur-driven Domain Adaptation}
While numerous domain adaptation works have been studied for other vision tasks~\cite{ AttentionDA, SigmaDA, CrossDA, MicDA, CatDA, Sharp2BlurHumanPoseDA, sida, emotion}, their application to motion deblurring has been relatively underexplored.
Among existing domain-adaptive deblurring methods, generating self-supervision signals from the test data has emerged as a dominant strategy. These methods generally fall into two categories: reconstruction-based and synthesis-based approaches.


Reconstruction-based approaches train models to reproduce the original blur during inference. Specifically, Chi \etal~\cite{MetaAux} and Nah \etal~\cite{CleanHard} proposed test-time adaptation schemes that derive supervision signals by reconstructing the input blurry images. However, this indirect supervision often provides insufficient guidance, restricting the model to a few update steps and yielding marginal performance gains.
Synthesis-based approaches provide more direct supervision by generating pseudo-training pairs. Recent works~\cite{metaTransfer, DADeblur} identify pseudo-sharp patches within the target domain and synthesize training pairs by blurring them with their inherent motion patterns. 
While effective for fine-tuning, these methods face the following limitations. 
First, they are constrained by the inherent entanglement of motion and content. Because motion is naturally embedded within consecutive frames, these methods can only re-apply motion to its original pseudo-sharp counterpart, limiting the combinatorial diversity required for adapting to various blurs in the target domain.
Second, relying on frame-based blurring makes them inapplicable to the well-developed event-based deblurring models~\cite{efnet, maenet, EIFNet, EBFI}.

To overcome these bottlenecks, we propose leveraging the high temporal resolution of events to explicitly decouple motion from visual content. This decoupling enables both intrinsic motion synthesis (using inherent motion) and extrinsic motion transfer across different sources, generating a diverse set of training pairs beyond naturally coupled combinations, while natively supporting event-based deblurring models.

\vspace{-3mm}
\section{Method}
\label{sec:method}

\begin{figure*}[t]
    \centering
    \includegraphics[width=0.95\textwidth]{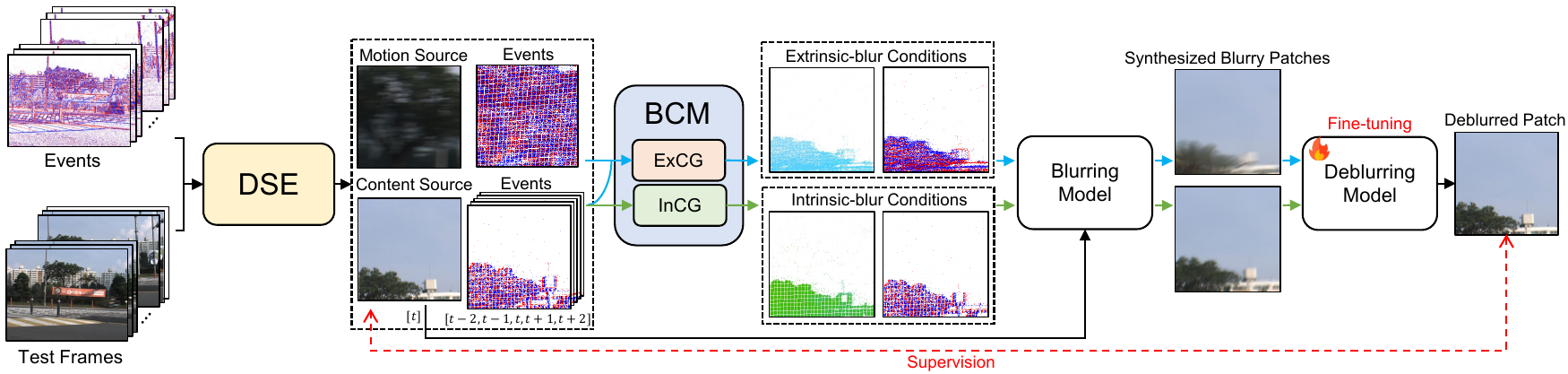}
    \vspace{-3mm}
    \caption{Overview of the proposed Event-guided Blur Synthesis (EvBS) framework. Given test frames and their corresponding event streams from the target domain, the Dual Source Extractor (DSE) extracts pseudo-sharp patches as content sources and blurred patches as motion sources. The Blur Conditioning Module (BCM) then generates diverse blur conditions via two pathways: the Intrinsic-blur Condition Generator (InCG) utilizes the inherent motion of the content sources, while the Extrinsic-blur Condition Generator (ExCG) transfers motions from the motion sources to different content sources. These conditions guide a blurring model to synthesize a diverse set of blurred patches. Finally, these synthesized patches are used to fine-tune a pre-trained deblurring model, with the original content sources serving as direct supervision (red dashed line).
    }
    \vspace{-3mm}
    \label{fig:overview}
\end{figure*}

\begin{figure}[t]
    \centering
    \includegraphics[width=0.48\textwidth]{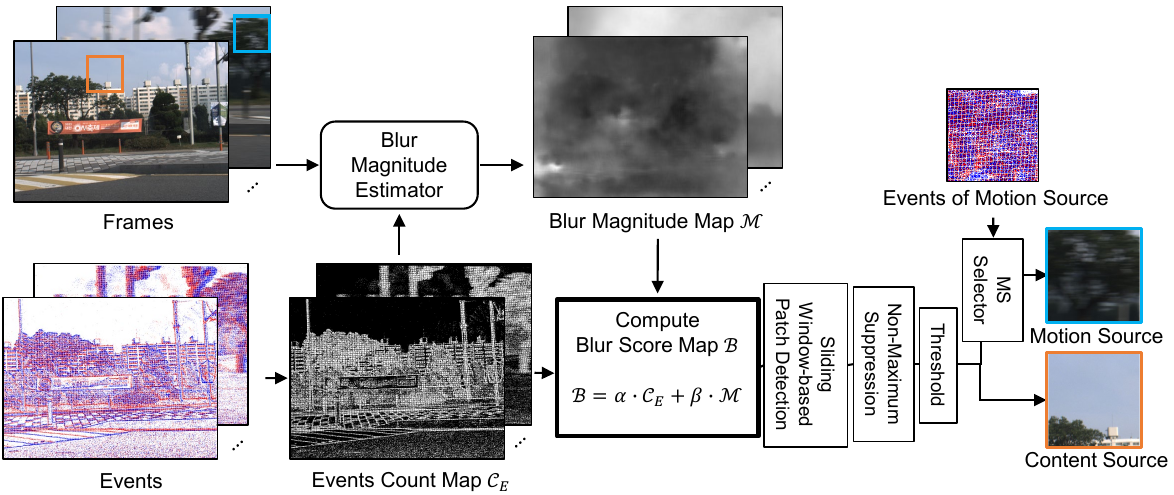}
    \caption{Illustration of the Dual Source Extractor. 
    The blur score map $\mathcal{B}$ is computed from the event count map $\mathcal{C}_E$ and the blur magnitude map $\mathcal{M}$. Based on this map, sliding-window detection identifies content sources (low scores) and motion source candidates (high scores). Subsequent non-maximum suppression (NMS) and thresholding steps are utilized to eliminate overlapping patches and reduce redundancy. Motion sources are further filtered by the motion source selector based on motion consistency.
    }
    \vspace{-3mm}
    \label{fig:dse}
\end{figure}

Due to the inherent entanglement between motion and content, existing blur-synthesis methods are limited to generating training pairs using only the motion patterns inherent to the sharp patches themselves. To overcome this, we propose leveraging events to explicitly decouple motion from visual content. This strategy allows us to synthesize diverse training pairs by utilizing both intrinsic motion (a content's own pattern) and extrinsic motion (patterns transferred from different sources).

As illustrated in Fig.~\ref{fig:overview}, our framework operates through two main components. First, the \textbf{Dual Source Extractor (DSE)} identifies and extracts pseudo-sharp content sources and distinct motion sources from the test data. Next, the \textbf{Blur Conditioning Module (BCM)} constructs diverse synthesis conditions through two pathways: the \textit{Intrinsic-blur Condition Generator (InCG)} applies a content source's own motion, whereas the \textit{Extrinsic-blur Condition Generator (ExCG)} transfers motion patterns from the motion sources to different content sources. These conditioning signals then guide a diffusion-based reblurring model~\cite{IDBlau} to synthesize the diverse training pairs required to fine-tune the pre-trained deblurring network on the target domain. The following subsections detail each framework component.

\vspace{-2mm}
\subsection{Dual Source Extractor (DSE)}
\label{ssec:dse}
To synthesize diverse training pairs for target domain adaptation, we first extract suitable content sources (pseudo-sharp patches) and motion sources (patches with clear motion patterns) from the given test frames. As shown in Fig.~\ref{fig:dse}, our DSE achieves this through a blur-aware patch detection approach.

\noindent \textbf{Blur score estimation.}
We leverage the observation that regions with high event activity typically correspond to areas with significant motion. From the input event stream, we generate an event count map $\mathcal{C}_E$ that accumulates the number of events at each pixel location. To complement this, we employ a Blur Magnitude Estimator (BME)~\cite{DADeblur}, which predicts the spatial distribution of blur intensity and produces a blur magnitude map $\mathcal{M}$.
Unlike the original BME that uses only RGB images, we enhance its estimation capability by retraining it with both the input image and the corresponding $\mathcal{C}_E$. This multi-modal integration allows the model to better capture motion-intensive regions.
We then compute a comprehensive Blur Score Map $\mathcal{B}$ by combining these two complementary cues:
\begin{equation}
\mathcal{B} = \alpha \cdot \mathcal{C}_E + \beta \cdot \mathcal{M},
\label{eq:blur_score}
\end{equation}
where $\alpha$ and $\beta$ are weighting parameters. Higher values in $\mathcal{B}$ indicate blurrier regions (motion source candidates), while lower values suggest sharper regions (content source candidates).

\noindent \textbf{Patch detection and refinement.}
Using the blur score map $\mathcal{B}$, we extract content and motion source candidates from regions where $\mathcal{B} < \tau$ and $\mathcal{B} > 1-\tau$, respectively. The threshold $\tau$ is adaptively adjusted~\cite{DADeblur} so that each source type constitutes approximately $r$\% of the total frames (ablated in Sec.~\ref{ssec:abs}). Following sliding-window detection and non-maximum suppression to remove overlapping patches, we further refine the motion source candidates. 
To ensure reliable motion transfer, we introduce a Motion Source (MS) Selector that evaluates the directional coherence of candidate patches. It explicitly filters out patches with chaotic or conflicting trajectories, retaining only those with highly coherent motions as the final sources.

\begin{figure}[t]
    \centering
    \includegraphics[width=0.44\textwidth]{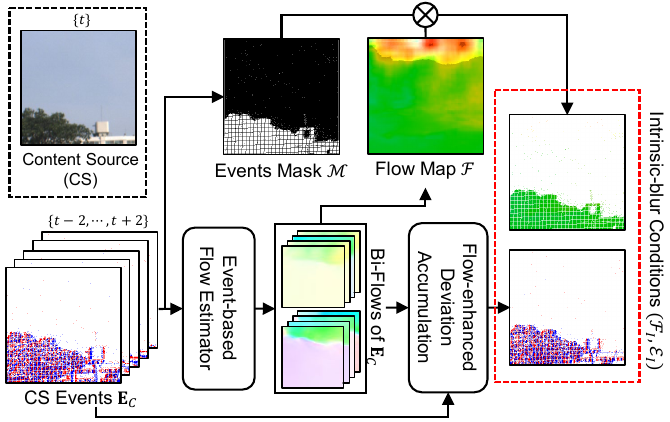}
    \caption{Intrinsic-blur Condition Generator. 
    Given events $\mathbf{E}_C$ from the temporal window $[t-2, t+2]$ around a content source at frame $t$, we first estimate bidirectional optical flows. These flows are aggregated into a flow map and then multiplied by an event mask $\mathcal{M}$ to produce the final intrinsic flow map $\mathcal{F}_I$. Concurrently, $\mathbf{E}_C$ is processed via Flow-Enhanced Deviation Accumulation (FEDA) to generate the intrinsic event representation $\mathcal{E}_I$. Together, the pair $(\mathcal{F}_I, \mathcal{E}_I)$ serves as the intrinsic-blur conditions.
    }
    \vspace{-3mm}
    \label{fig:incg}
\end{figure}

\begin{figure}[t]
    \centering
    \includegraphics[width=0.5\textwidth]{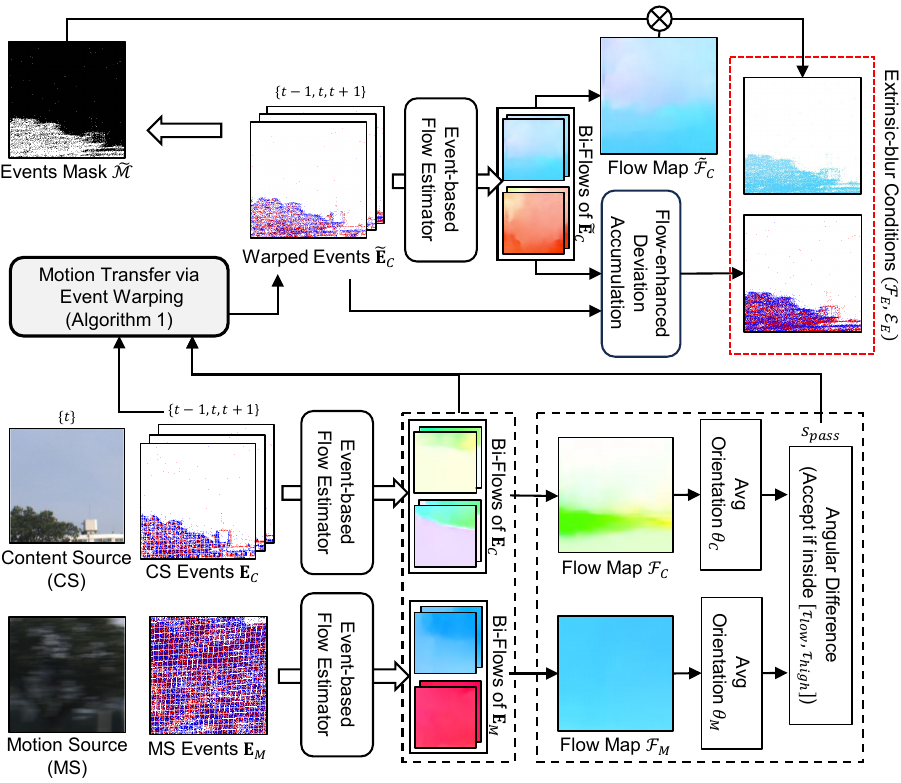}
    \vspace{-3mm}
    \caption{
    Extrinsic-blur Condition Generator. 
    Flow maps $\mathcal{F}_C$ and $\mathcal{F}_M$ are first extracted from content events $\mathbf{E}_C$ and motion events $\mathbf{E}_M$. If their angular difference falls within $[\tau_{low}, \tau_{high}]$, the motion pattern from $\mathbf{E}_M$ is transferred to $\mathbf{E}_C$ via Algorithm~\ref{alg:motion_transfer} to produce the warped events $\tilde{\mathbf{E}}_C$. Finally, $\tilde{\mathbf{E}}_C$ is processed via FEDA to yield the extrinsic flow map $\mathcal{F}_E$ and event representation $\mathcal{E}_E$. The resulting condition $(\mathcal{F}_E, \mathcal{E}_E)$ effectively integrates the content's spatial structure with external motion patterns.
    }
    \vspace{-3mm}
    \label{fig:excg}
\end{figure}

\vspace{-3mm}
\subsection{Blur Conditioning Module (BCM)}
\label{ssec:mcm}
Once content and motion sources are extracted, the Blur Conditioning Module (BCM) generates multi-modal motion conditions to guide the subsequent blur synthesis. For the synthesis process, we build upon ID-Blau~\cite{IDBlau}. Let $\mathcal{D}_\psi$ denote this conditional diffusion model parameterized by $\psi$, which was originally designed to generate a blurred image $B$ given a sharp content image $C$ and a flow map $\mathcal{F} = [\tilde{u}; \tilde{v}; m] \in \mathbb{R}^{3 \times H \times W}$ (encoding unit orientation $(\tilde{u}, \tilde{v})$ and magnitude $m$):
\begin{equation}
B = \mathcal{D}_\psi(\mathcal{F}, C).
\end{equation}

While the flow map $\mathcal{F}$ effectively captures macro-level motion trajectories, it lacks the temporal granularity required to model local micro-motion dynamics.
To overcome this, our BCM leverages event data as an auxiliary conditioning signal. Specifically, we introduce Flow-Enhanced Deviation Accumulation (FEDA) to supplement the macro motion of $\mathcal{F}$ with the fine-grained, high-frequency dynamics embedded in events

The deviation accumulation representation~\cite{maenet} effectively captures pure motion contours by aggregating the temporal variations of event streams. Building upon this, FEDA advances the representation by weighting the events with the magnitude derived from the corresponding motion vector. This approach preserves the overall motion trajectory while emphasizing regions with high motion intensity (detailed in the Supplementary Material). Given this event-guided FEDA representation $\mathcal{E} \in \mathbb{R}^{H \times W}$, we train a modified $\mathcal{D}_\psi$ from scratch to synthesize the final blurred image $B$:
\begin{equation}
B =\mathcal{D}_\psi(\mathcal{F}, \mathcal{E}, C).
\end{equation}
This refined, event-guided conditioning improves the domain-specific blur synthesis, ultimately boosting the efficacy of the fine-tuning process (see Sec.~\ref{ssec:abs}).

To formulate the synthesis conditions ($\mathcal{F}$ and $\mathcal{E}$), the BCM comprises two complementary generators: the Intrinsic-blur Condition Generator (InCG) and the Extrinsic-blur Condition Generator (ExCG). 
Both generators operate on the shared principle of explicitly decoupling motion from visual content by extracting motion cues exclusively from event streams.
Building upon this decoupled cues, InCG constructs conditioning signals using the content source's own motion patterns. In contrast, ExCG formulates conditions by transferring the motion patterns of an external motion source to a distinct content source.

\noindent \textbf{Intrinsic-blur Condition Generator (InCG).}
InCG generates conditions for blurring each content source using its own motion pattern, as illustrated in Fig.~\ref{fig:incg}. Given a content source from the $t$-th frame, we utilize the corresponding events $\mathbf{E}_C$ from the temporal window $[t-2, t+2]$ to capture the local motion dynamics. These events are fed into an event-based optical flow estimator~\cite{ERAFT} to extract forward flows $\{\mathbf{F}_{t-2 \rightarrow t-1}, \ldots, \mathbf{F}_{t+1 \rightarrow t+2}\}$ and backward flows $\{\mathbf{F}_{t-1 \rightarrow t-2}, \ldots, \mathbf{F}_{t+2 \rightarrow t+1}\}$.

We then compute a flow map by aggregating these bidirectional flows~\cite{IDBlau}:
$$\mathcal{F} = \sum_{n=t-2}^{t+1} \frac{\mathbf{F}_{n \rightarrow n+1} - \mathbf{F}_{n+1 \rightarrow n}}{2},$$ where $\mathcal{F} = [u; v] \in \mathbb{R}^{2 \times H \times W}$ represents the horizontal and vertical motion components. For conditioning, we reformulate this into a 3-channel tensor $\mathcal{F} = [\tilde{u}; \tilde{v}; \tilde{m}] \in \mathbb{R}^{3 \times H \times W}$, where $\tilde{u}$ and $\tilde{v}$ encode the normalized motion orientation, and $\tilde{m}$ encodes the normalized magnitude:
\begin{equation}
\tilde{u} = \frac{u}{\sqrt{u^2 + v^2}}, \quad 
\tilde{v} = \frac{v}{\sqrt{u^2 + v^2}}, \quad 
\tilde{m} = \frac{\sqrt{u^2 + v^2}}{M},
\label{eq:flow_norm}
\end{equation}
where $M$ denotes the maximum magnitude within the current map.
To focus exclusively on motion-active regions, we create a binary mask $\mathcal{M} \in \{0,1\}^{H \times W}$, where $\mathcal{M}(x,y) = 1$ if at least one event occurred at $(x,y)$, and $0$ otherwise. We apply this mask to obtain the final intrinsic flow map $\mathcal{F}_I = \mathcal{F} \odot \mathcal{M}$, where $\odot$ denotes element-wise multiplication. Concurrently, we process $\mathbf{E}_C$ via the FEDA algorithm to generate the intrinsic event representation $\mathcal{E}_I \in \mathbb{R}^{H \times W}$.

Consequently, the pair $(\mathcal{F}_I, \mathcal{E}_I)$ serves as the intrinsic-blur condition, effectively guiding the blurring model to synthesize blur that is governed by the content source's inherent motion.

\noindent \textbf{Extrinsic-blur Condition Generator (ExCG).}
As illustrated in Fig.~\ref{fig:excg}, the ExCG formulates conditioning signals by transferring extrinsic motion patterns from distinct motion sources to a target content source.
Given a content source from the $t$-th frame, we extract corresponding events from the narrower temporal window $[t-1, t+1]$ to minimize misalignment during motion transfer. 
Using an event-based flow estimator~\cite{ERAFT}, we compute the bidirectional flows and aggregate them into a content flow map $\mathcal{F}_C = \sum_{n=t-1}^{t} (\mathbf{F}_{n \rightarrow n+1} - \mathbf{F}_{n+1 \rightarrow n})/2$.
Concurrently, for a given motion source spanning the temporal interval $[t_s, t_e]$, we divide its continuous event stream into three consecutive temporal segments. We then extract the bidirectional flows between these sequential segments and aggregate them to construct a motion flow map $\mathcal{F}_M = \sum_{i=1}^{2} (\mathbf{F}_{i \rightarrow i+1} - \mathbf{F}_{i+1 \rightarrow i})/2$, where $i$ denotes the segment indices. Here, both flow maps $\mathcal{F}_C, \mathcal{F}_M \in \mathbb{R}^{2 \times H \times W}$ encode the pixel-wise motion vectors $[u; v]$.

To identify motion sources that offer sufficient diversity while remaining transferable, we evaluate their orientation similarity with the target content source. 
First, we compute the average orientation vectors $\boldsymbol{\theta}_C, \boldsymbol{\theta}_M \in \mathbb{R}^2$ from $\mathcal{F}_C$ and $\mathcal{F}_M$. For a given flow map, the average components are calculated as:
\begin{equation}
\bar{u} = \frac{1}{N}\sum u, \quad
\bar{v} = \frac{1}{N}\sum v, \quad
\boldsymbol{\theta} = \frac{[\bar{u}; \bar{v}]}{\|[\bar{u}; \bar{v}]\|},
\label{eq:flow_avg}
\end{equation}
where the summation is computed over all valid pixels in the flow map, and $N$ denotes the total number of such pixels.
We then compute the angular difference $\phi = \arccos(\boldsymbol{\theta}_C \cdot \boldsymbol{\theta}_M)$ between the content orientation and motion orientation. We proceed with the motion transfer only if this angle falls within a specific range, $\tau_{low} < \phi < \tau_{high}$ (ablated in Sec.~\ref{ssec:abs}). This filtering excludes redundant motion sources that may limit diversity and highly disparate ones that cause unrealistic synthesized blur.

\begin{algorithm}[t]
\caption{Motion Transfer via Event Warping}
\label{alg:motion_transfer}
\begin{algorithmic}[1]
\STATE \textbf{Input:} \\
                        Content source events $\mathbf{E}_C$,
                        \\
                        Bi-Flows of $\mathbf{E}_C$: $\{\mathbf{F}_{t-1 \rightarrow t}, \mathbf{F}_{t \rightarrow t+1}\}$,
                       $\{\mathbf{F}_{t \rightarrow t-1}, \mathbf{F}_{t+1 \rightarrow t}\}$.
                       \\
                       Bi-Flows of $\mathbf{E}_M$: $\{\mathbf{F}_{i-1 \rightarrow i}, \mathbf{F}_{i \rightarrow i+1}\}, \{\mathbf{F}_{i \rightarrow i-1}, \mathbf{F}_{i+1 \rightarrow i}\}$.                    
                      \\
                      Similarity check result $s_{\text{pass}} \in \{\text{true}, \text{false}\}$
\STATE \textbf{Output:} Warped events $\tilde{\mathbf{E}}_C$

\IF{$s_{\text{pass}} = \text{false}$}
    \STATE \textbf{return} $\emptyset$ 
\ENDIF

\STATE Get median time $t_{\text{med}}$ from $\mathbf{E}_C$ at $t$

\FOR{each direction $k \in \{1, -1\}$}
    \STATE \textit{// Get directional flows}
    \STATE $\mathcal{F}_C^k \gets (\mathbf{F}_{t \rightarrow t+k} - \mathbf{F}_{t+k \rightarrow t})/2$
    \STATE $\mathcal{F}_M^k \gets (\mathbf{F}_{i \rightarrow i+k} - \mathbf{F}_{i+k \rightarrow i})/2$
    
    \STATE \textit{// Extract local \& global motion (Eq.~\ref{eq:local_global})}
    \STATE $\mathcal{F}_C^k \rightarrow \mathbf{V}_C^k$, $\mathcal{F}_M^k \rightarrow \mathbf{v}_M^k$
    
    \STATE \textit{// Compute offset field}
    \STATE $\mathbf{T}_k \gets \mathbf{v}_M^k - \mathbf{V}_C^k$
    
    \STATE \textit{// Split events $\mathbf{E}_C$}
    \STATE $\mathbf{E}_k \gets$ $\mathbf{E}_C$ in $[t_{\text{med}}, t+1]$ if $k=1$, else $[t-1, t_{\text{med}}]$
    \STATE \textit{// Set bounds}
    \STATE $[t_{\min}, t_{\max}] \gets [t_{\text{med}}, t+1]$ if $k=1$, else $[t-1, t_{\text{med}}]$
    
    \FOR{each event $(x, y, \tau) \in \mathbf{E}_k$}
        \STATE $w(\tau) \gets \frac{1 - k}{2} + k \cdot \frac{\tau - t_{\min}}{t_{\max} - t_{\min}}$
        \STATE $(x', y') \gets (x, y) + w(\tau) \cdot \mathbf{T}_k(x,y)$
        \STATE Add $(x', y', \tau)$ to $\tilde{\mathbf{E}}_C$
    \ENDFOR
\ENDFOR
\STATE \textbf{return} Warped events $\tilde{\mathbf{E}}_C$

\end{algorithmic}
\end{algorithm}

Once a motion source passes the similarity check, we transfer its motion pattern to the content source via event warping, as detailed in Algorithm~\ref{alg:motion_transfer}. 
The core idea is to compute an offset field $\mathbf{T}_k(x,y)$ that represents the difference between the target motion source's global motion and the original content source's local motion.

Specifically, for each temporal direction ($k \in \{1, -1\}$), we first compute the bidirectionally aggregated flow maps $\mathcal{F}_C^k=[u_c; v_c]$ and $\mathcal{F}_M^k=[u_m;v_m] \in \mathbb{R}^{2 \times H \times W}$.  
For the content source, we directly use the flow map $\mathcal{F}_C^k$ as the local motion map $\mathbf{V}_C^k$.
To extract the global motion vector $\mathbf{v}_M^k$ from $\mathcal{F}_M^k$, we spatially average its components across $N$ valid pixels: $\bar{u}_m = \frac{1}{N}\sum u_m$, $\bar{v}_m = \frac{1}{N}\sum v_m$, and $\bar{m} = \frac{1}{N}\sum m$, where $m = \sqrt{{u_m}^2 + {v_m}^2}$. 
Then, the local motion map $\mathbf{V}_C^k \in \mathbb{R}^{2 \times H \times W}$ and the global motion vector $\mathbf{v}_M^k \in \mathbb{R}^2$ are formally defined as:
\begin{equation}
       \mathbf{V}_C^k=[u_c; v_c] ,\quad 
       \mathbf{v}_M^k=\frac{[\bar{u}_m; \bar{v}_m]}{\|[\bar{u}_m; \bar{v}_m]\|} \cdot \bar{m}.
\label{eq:local_global}
\end{equation}

We then compute the offset field $\mathbf{T}_k \in \mathbb{R}^{2 \times H \times W}$ by subtracting the local motion map from the spatially broadcasted global motion vector:
\begin{equation}
\mathbf{T}_k = \mathbf{v}_M^k - \mathbf{V}_C^k.
\label{eq:offset}
\end{equation}
Finally, each event in $\mathbf{E}_C$ at timestamp $\tau$ is warped using $\mathbf{T}_k$, scaled by a temporal weight $w(\tau)$ proportional to its distance from the $\mathbf{E}_C$'s median time $t_{\text{med}}$.
This operation effectively neutralizes the original inherent motion pattern in $\mathbf{E}_C$ while injecting the target extrinsic pattern, resulting in the warped events $\tilde{\mathbf{E}}_C$.

Using the warped events $\tilde{\mathbf{E}}_C$, we extract new bidirectional flows $\tilde{\mathbf{F}}$ and a corresponding event mask $\tilde{\mathcal{M}}$ to compute the extrinsic flow map $\mathcal{F}_E = \big(\sum_{n=t-1}^{t} (\tilde{\mathbf{F}}_{n \rightarrow n+1} - \tilde{\mathbf{F}}_{n+1 \rightarrow n})/2\big) \odot \tilde{\mathcal{M}}$. 
Concurrently, $\tilde{\mathbf{E}}_C$ is processed via FEDA to yield the extrinsic event representation $\mathcal{E}_E$. Ultimately, the pair $(\mathcal{F}_E, \mathcal{E}_E)$ serves as the extrinsic-blur condition, guiding the model to synthesize blur with the transferred motion pattern.

\begin{table*}[!t]
\centering
\caption{Quantitative results across deblurring benchmarks. We compare baseline performance with results obtained after adaptation via our method (w/ Ours). * indicates event-based models.}
\vspace{-3mm}
\label{tab:quan_results}
\resizebox{0.9\textwidth}{!}{
\begin{tabular}{l|ccc|ccc|ccc|ccc|ccc}
\toprule
\multirow{2}{*}{Models} & \multicolumn{3}{c|}{EVRB} & \multicolumn{3}{c|}{REVD} & \multicolumn{3}{c|}{HighREV-3} & \multicolumn{3}{c|}{HighREV-7} & \multicolumn{3}{c}{HighREV-11} \\
& PSNR$\uparrow$ & SSIM$\uparrow$ & LPIPS$\downarrow$ & PSNR$\uparrow$ & SSIM$\uparrow$ & LPIPS$\downarrow$ & PSNR$\uparrow$ & SSIM$\uparrow$ & LPIPS$\downarrow$ & PSNR$\uparrow$ & SSIM$\uparrow$ & LPIPS$\downarrow$ & PSNR$\uparrow$ & SSIM$\uparrow$ & LPIPS$\downarrow$ \\
\midrule
MAENet$^*$ & 25.46 & 0.791 & 0.428 & 25.41 & 0.813 & 0.421 & 36.35 & 0.939 & 0.303 & 31.79 & 0.859 & 0.366 & 27.24 & 0.771 & 0.447 \\
\quad w/ Ours & 27.93 & 0.851 & 0.386 & 28.64 & 0.858 & 0.372 & 39.66 & 0.965 & 0.245 & 33.91 & 0.929 & 0.309 & 29.92 & 0.832 & 0.391 \\
\midrule
EFNet$^*$ & 25.77 & 0.797 & 0.419 & 26.39 & 0.827 & 0.394 & 37.34 & 0.942 & 0.293 & 32.35 & 0.873 & 0.359 & 27.77 & 0.791 & 0.432 \\
\quad w/ Ours & 28.17 & 0.859 & 0.375 & 28.96 & 0.865 & 0.363 & 40.23 & 0.977 & 0.212 & 34.51 & 0.942 & 0.305 & 30.01 & 0.844 & 0.379 \\
\midrule
EBFI$^*$ & 25.31 & 0.785 & 0.435 & 26.14 & 0.829 & 0.402 & 38.22 & 0.943 & 0.299 & 31.93 & 0.860 & 0.377 & 27.33 & 0.778 & 0.442 \\
\quad w/ Ours & 27.51 & 0.845 & 0.389 & 28.46 & 0.862 & 0.357 & 40.67 & 0.970 & 0.242 & 34.65 & 0.941 & 0.309 & 29.85 & 0.835 & 0.397 \\
\midrule
NAFNet & 24.13 & 0.746 & 0.473 & 24.74 & 0.794 & 0.452 & 36.16 & 0.917 & 0.322 & 29.76 & 0.828 & 0.415 & 25.57 & 0.738 & 0.486 \\
\quad w/ Ours & 26.24 & 0.817 & 0.419 & 26.55 & 0.838 & 0.399 & 38.53 & 0.961 & 0.268 & 31.95 & 0.861 & 0.358 & 27.63 & 0.784 & 0.426 \\
\midrule
MIMO-UNet & 23.97 & 0.732 & 0.487 & 24.34 & 0.789 & 0.459 & 36.69 & 0.922 & 0.326 & 29.79 & 0.835 & 0.406 & 25.01 & 0.727 & 0.491 \\
\quad w/ Ours & 26.01 & 0.807 & 0.423 & 26.18 & 0.825 & 0.401 & 38.11 & 0.959 & 0.277 & 32.03 & 0.871 & 0.349 & 27.23 & 0.772 & 0.431 \\
\bottomrule
\end{tabular}
}
\vspace{-3mm}
\end{table*}

\begin{table}[t]
\centering
\caption{Comparison with recent domain-adaptive deblurring methods. Best results in \textbf{bold}.}
\label{tab:comparison_results}
\resizebox{\columnwidth}{!}{
\begin{tabular}{l|cc|cc|cc|cc|cc}
\toprule
\multirow{2}{*}{Models} & \multicolumn{2}{c|}{EVRB} & \multicolumn{2}{c|}{REVD} & \multicolumn{2}{c|}{HighREV-3} & \multicolumn{2}{c|}{HighREV-7} & \multicolumn{2}{c}{HighREV-11} \\
& PSNR & SSIM & PSNR & SSIM & PSNR & SSIM & PSNR & SSIM & PSNR & SSIM \\
\midrule
NAFNet & 24.13 & .746 & 24.74 & .794 & 36.16 & .917 & 29.76 & .828 & 25.57 & .738 \\
+ MetaDeblur & 24.22 & .753 & 24.81 & .797 & 36.54 & .929 & 29.97 & .832 & 25.60 & .739 \\
+ Blur2Blur & 24.72 & .770   & 25.59 & .809 & 37.28 & .937 & 30.38 & .838 & 26.25 & .750 \\
+ DADeblur & 25.23 & .781 & 25.87 & .819 & 37.71 & .946 & 31.28 & .849 & 27.19 & .767 \\
+ Ours & \textbf{26.24} & \textbf{.817} & \textbf{26.55} & \textbf{.838} & \textbf{38.53} & \textbf{.961} & \textbf{31.95} & \textbf{.861} & \textbf{27.63} & \textbf{.784} \\
\midrule
MIMO-UNet & 23.97 & .732 & 24.34 & .789 & 36.69 & .922 & 29.79 & .835 & 25.01 & .727 \\
+ MetaDeblur & 24.05 & .736 & 24.43 & .795 & 36.87 & .935 & 29.92 & .842 & 25.02 & .727 \\
+ Blur2Blur & 24.49 & .752 & 25.07 & .802 & 37.52 & .941 & 30.24 & .846 & 25.71 & .739 \\
+ DADeblur & 25.18 & .774 & 25.63 & .811 & 37.77 & .943 & 31.09 & .856 &26.87 & .755 \\
+ Ours & \textbf{26.01} & \textbf{.807} & \textbf{26.18} & \textbf{.825} & \textbf{38.11} & \textbf{.959} & \textbf{32.03} & \textbf{.871} & \textbf{27.23} & \textbf{.772} \\
\bottomrule
\end{tabular}
}
\vspace{-3mm}
\end{table}

\subsection{Blur Synthesis}
\label{ssec:pair_synthesis}
The generated intrinsic and extrinsic conditions are fed into the modified ID-Blau $\mathcal{D}_\psi$ to synthesize the final blurred images $B_t^I$ and $B_t^E$ for a given content source $C_t$:
\begin{equation}
B_t^I = \mathcal{D}_\psi(\mathcal{F}_I, \mathcal{E}_I, C_t), \quad B_t^E = \mathcal{D}_\psi(\mathcal{F}_E, \mathcal{E}_E, C_t).
\end{equation}
This process yields diverse blur-sharp training pairs tailored for target domain adaptation.
Specifically, when fine-tuning an event-based deblurring model, it takes the synthesized blurred patch (either $B_t^I$ or $B_t^E$) alongside its corresponding events (either $\mathbf{E}_C$ or $\tilde{\mathbf{E}}_C$) as inputs to restore the original pseudo-sharp content $C_t$.

\noindent \textbf{Extrinsic pair selection.} When multiple motion sources satisfy the orientation criterion, we prioritize the top-$n$ pairs that best preserve motion consistency. To quantify this, we compute a motion consistency score $\text{MCS} = (\boldsymbol{\tilde{\theta}}_C \cdot \boldsymbol{\theta}_M + 1) / 2$ between the average orientation vectors $\boldsymbol{\tilde{\theta}}_C, \boldsymbol{\theta}_M \in \mathbb{R}^2$ derived from $\tilde{\mathcal{F}}_C$ and $\mathcal{F}_M$ via Eq.~\ref{eq:flow_avg}. We then select the top-$n$ motion sources with the highest MCS, where $n$ is determined through ablation studies (Sec.~\ref{ssec:abs}).


\vspace{-2mm}
\section{Experiments}
\label{sec:experiments}
\subsection{Experimental Setups}
\label{ssec:setup}
\noindent \textbf{Datasets.}
We retrain the Blur Magnitude Estimator (BME)~\cite{DADeblur} and ID-Blau~\cite{IDBlau} on the GoPro~\cite{gopro} training set with synthetic events~\cite{esim} using hot-pixel noise and a Gaussian-distributed contrast threshold. 
For evaluation, we employ benchmarks covering both real and synthesized event-based deblurring datasets. Real-world performance is assessed on EVRB~\cite{CMTA} and REVD~\cite{FEVD}, which provide real motion blur and events from hybrid cameras. To overcome the limited availability of real-world event-based deblurring datasets, we utilize HighREV~\cite{refid}. Using the realistic blur synthesis pipeline in~\cite{RealBlurSynthesis}, we generate three variants — HighREV-3, HighREV-7, and HighREV-11 — by aggregating 3, 7, and 11 consecutive frames, respectively, to represent a comprehensive spectrum of blur intensities.

\noindent\textbf{Implementation Details.}
We employ E-RAFT~\cite{ERAFT} for event-based flow estimation. For BME~\cite{DADeblur} and ID-Blau~\cite{IDBlau}, we follow their original training settings while modifying their input spaces: BME is adapted to accept both RGB images and event count maps $\mathcal{C}_E$, and ID-Blau is updated to take the flow map $\mathcal{F}$, FEDA representation $\mathcal{E}$, and content image $C$ as conditions.

We evaluate our framework by fine-tuning five recent GoPro-pretrained deblurring models, including event-based (EFNet \cite{efnet}, MAENet~\cite{maenet}, EBFI~\cite{EBFI}) and frame-only (NAFNet~\cite{nafnet}, MIMO-UNet~\cite{mimounet}) architectures. All models are fine-tuned for 10 epochs on our synthesized pairs using the Charbonnier loss~\cite{char} and Adam optimizer~\cite{adam}, with the learning rate cosine-annealed~\cite{cosine_anneal} from $10^{-5}$ to $10^{-7}$. We report PSNR, SSIM~\cite{ssim}, and LPIPS~\cite{lpips}.
Key hyperparameters (ablated in Sec.~\ref{ssec:abs}) include the source selection ratio $r=0.15$, the number of extrinsic pairs $n=2$, and orientation angle thresholds $[\tau_{\text{low}}, \tau_{\text{high}}] = [15^\circ, 30^\circ]$. Empirical settings include: a patch size of $256 \times 256$  with a stride of $64$ for sliding window-based patch detection, initial patch threshold $\tau=0.05$, NMS IoU threshold of 0.3, and blur score weights $\alpha=\beta=0.5$ (Eq.~\ref{eq:blur_score}).

\begin{figure*}[t]
    \centering
    \begin{subfigure}[b]{0.49\textwidth}
        \centering
        \includegraphics[width=0.95\textwidth]{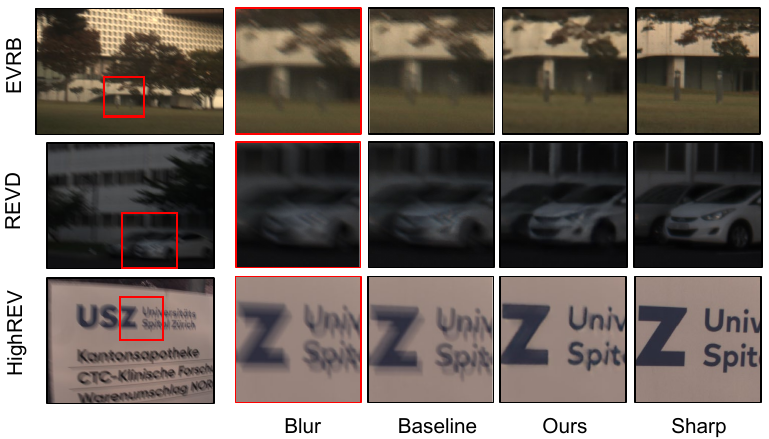} 
        \vspace{-1mm}
        \caption{MAENet}
        \label{fig:6a}
    \end{subfigure}
    \begin{subfigure}[b]{0.49\textwidth}
        \centering
        \includegraphics[width=0.95\textwidth]{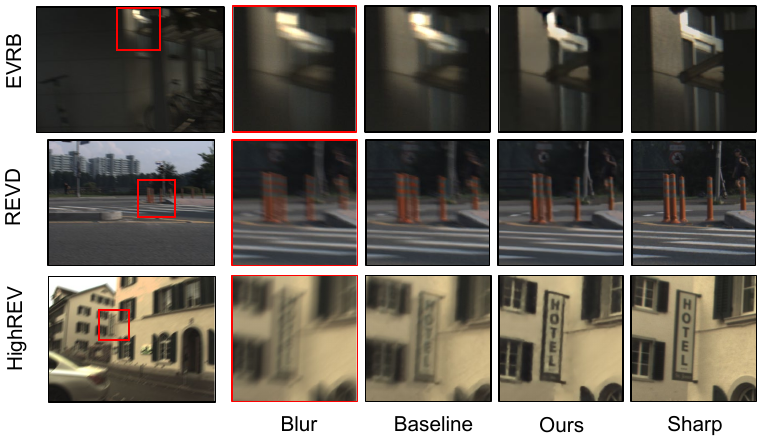} 
        \vspace{-1mm}
        \caption{EFNet}
        \label{fig:6b}
    \end{subfigure}

    \vspace{-3mm}
    \caption{
    Qualitative results across diverse models and datasets. 
    We compare the blurry input (blur), pre-trained baseline (baseline), our adapted model (ours), and ground truth (sharp), with red boxes denoting zoomed-in regions.
    Our approach consistently restores finer details across both event-based deblurring models. Additional visual results for the other models (EBFI, NAFNet, MIMO-UNet) are shown in the supplementary material.
    }
    \vspace{-3mm}
    \label{fig:deblur_results}
\end{figure*}

\begin{figure}[t]
    \centering
    \includegraphics[width=0.5\textwidth]{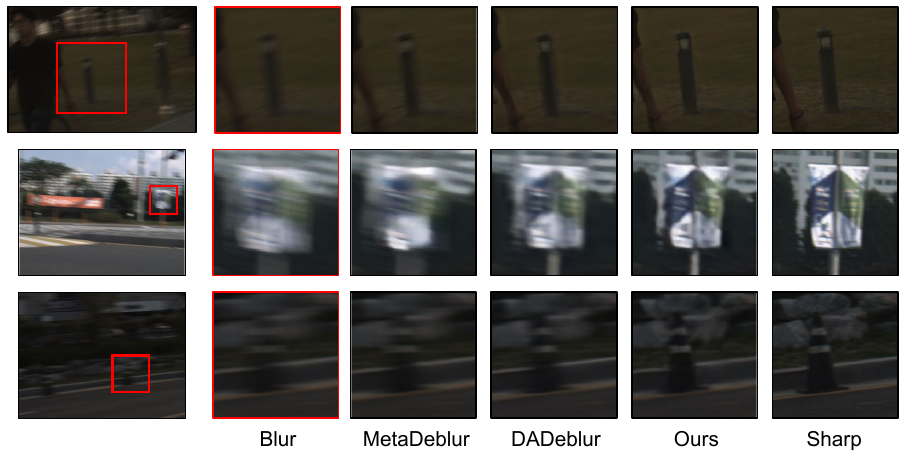}
    \vspace{-3mm}
    \caption{Visual comparison of deblurring results against recent domain-adaptive methods~\cite{metaTransfer, DADeblur} Our method consistently yields sharper and more visually pleasing results across diverse scenes.}
    \vspace{-3mm}
    \label{fig:visual_comparison}
\end{figure}

\subsection{Experimental Results}
\label{ssec:result}
\noindent\textbf{Quantitative Results.}
Table~\ref{tab:quan_results} presents the quantitative evaluation on various deblurring benchmarks. Our method consistently boosts performance across diverse architectures, including event-based (MAENet, EFNet, EBFI) and frame-only models (NAFNet, MIMO-UNet). Specifically, we achieve substantial average gains of 2.24 dB on EVRB, 2.35 dB on REVD, and up to 2.63 dB on HighREV datasets. 

\begin{figure}[t]
    \centering
    \includegraphics[width=0.48\textwidth]{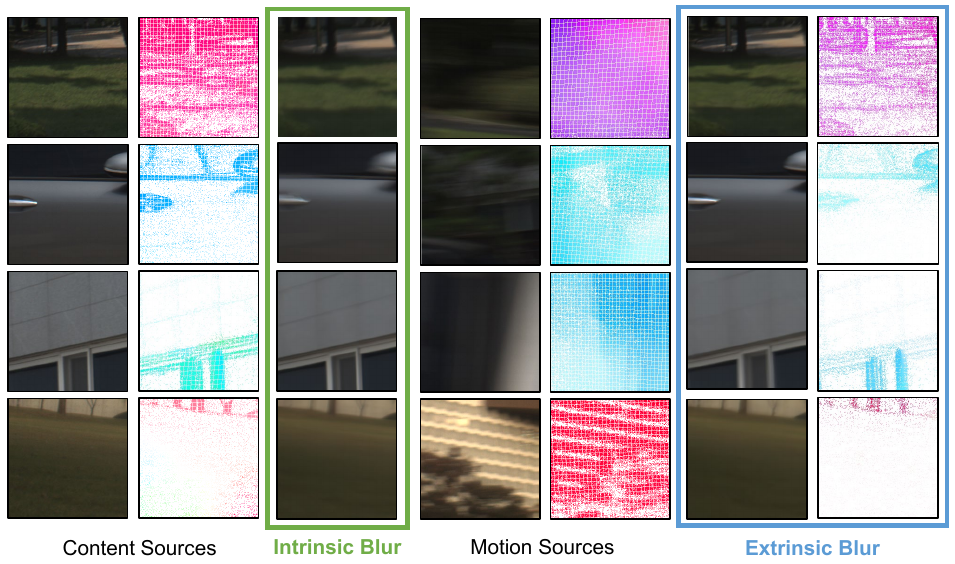}
    \vspace{-3mm}
    \caption{
    Qualitative examples of intrinsic and extrinsic blur synthesis. For each row, we display the content source and its flow map $\mathcal{F}_C$ (left), followed by the \textcolor{LimeGreen}{intrinsic synthesized blur}. To demonstrate motion transfer, we show a separate motion source and its flow map $\mathcal{F}_M$ (middle), and the resulting \textcolor{RoyalBlue}{extrinsic synthesized blur} with the transferred flow map $\tilde{\mathcal{F}}_C$ (right). All flow maps are masked by event masks to explicitly highlight motion-intensive regions.
    }
    \vspace{-3mm}
    \label{fig:motion_synthesis}
\end{figure}

Table~\ref{tab:comparison_results} compares our method with recent domain-adaptive deblurring approaches~\cite{metaTransfer, DADeblur, Blur2Blur}. In addition to synthesis-based methods~\cite{metaTransfer, DADeblur}, we include Blur2Blur~\cite{Blur2Blur}, which adapts unseen target blur to blur patterns observed during training. As these methods cannot process event streams, we conduct the comparison using frame-only backbones (NAFNet and MIMO-UNet). EvBS consistently achieves the best performance, with gains of up to 1.01\,dB on NAFNet and 0.94\,dB on MIMO-UNet compared with the strongest baselines.

\begin{figure*}[t]
    \centering
    \begin{subfigure}[b]{0.32\textwidth}
        \centering
        \includegraphics[width=0.96\textwidth]{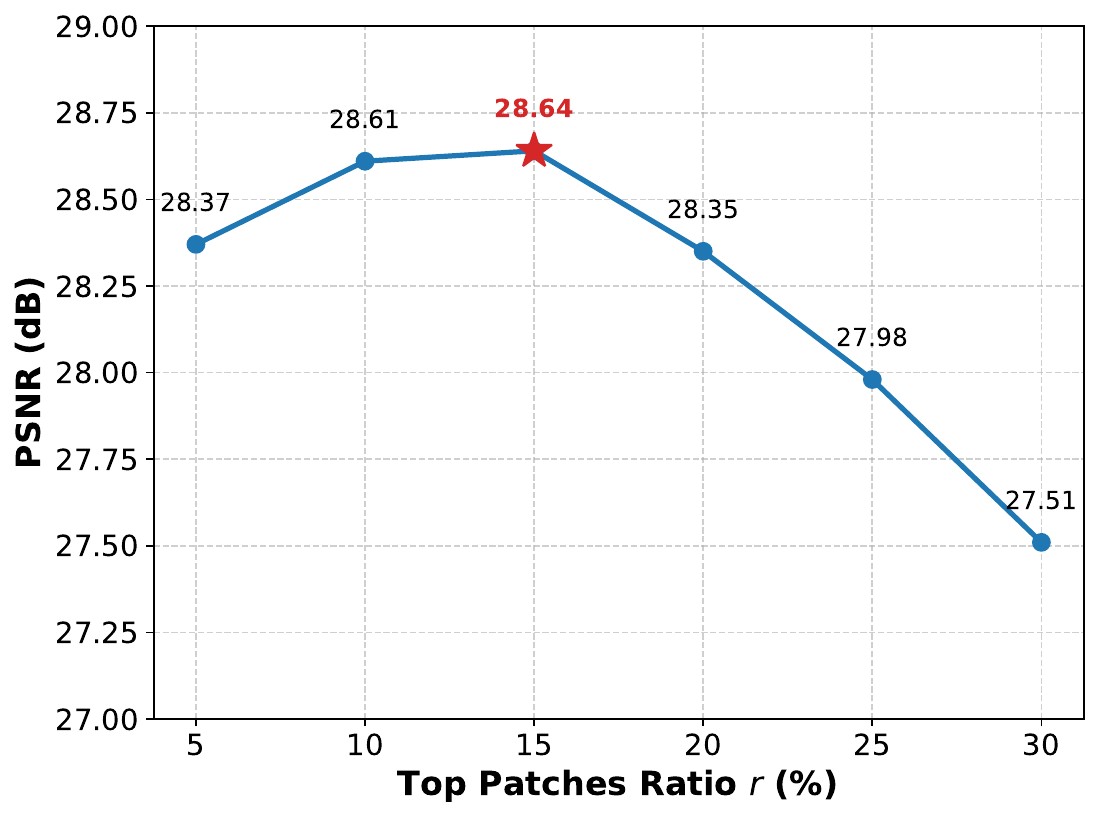} 
        \vspace{-1.5mm}
        \caption{}
        \label{fig:8a}
    \end{subfigure}
    \begin{subfigure}[b]{0.32\textwidth}
        \centering
        \includegraphics[width=0.96\textwidth]{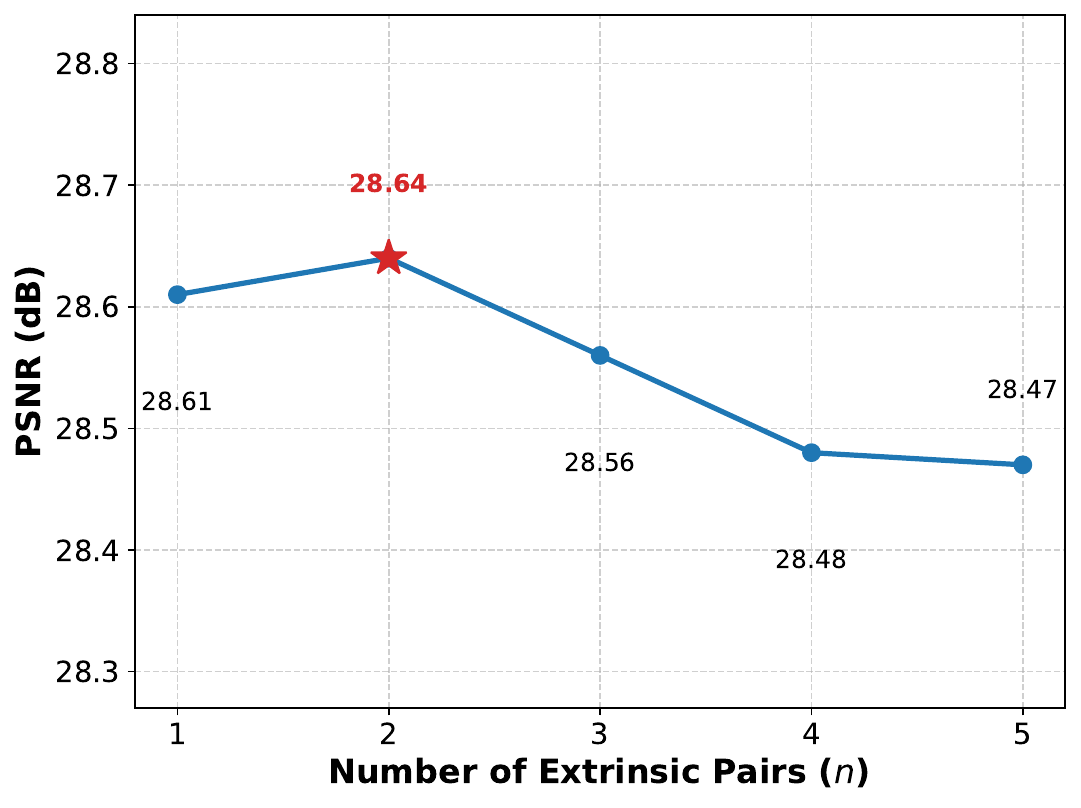} 
        \vspace{-1.5mm}
        \caption{}
        \label{fig:8b}
    \end{subfigure}
    \begin{subfigure}[b]{0.32\textwidth}
        \centering
        \includegraphics[width=0.96\textwidth]{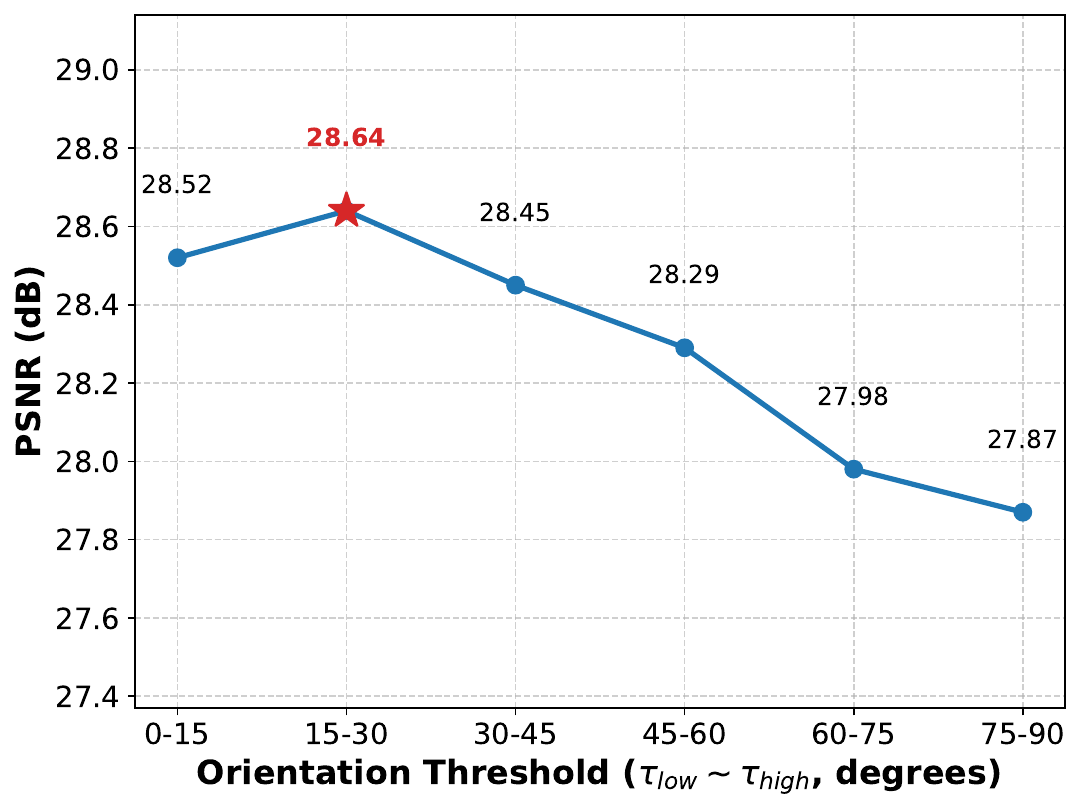} 
        \vspace{-1.5mm}
        \caption{}
        \label{fig:8c}
    \end{subfigure}

    \vspace{-3mm}
    \caption{
    Ablation study on key hyperparameters. We analyze the impact of (a) the top patches ratio $r$, (b) the number of extrinsic pairs $n$, and (c) the orientation threshold ranges $[\tau_{low}, \tau_{high}]$. The best values are marked with red stars.
    }
    \label{fig:ablation_hyperparams}
\end{figure*}

\begin{figure}[t]
    \centering
    \includegraphics[width=0.5\textwidth]{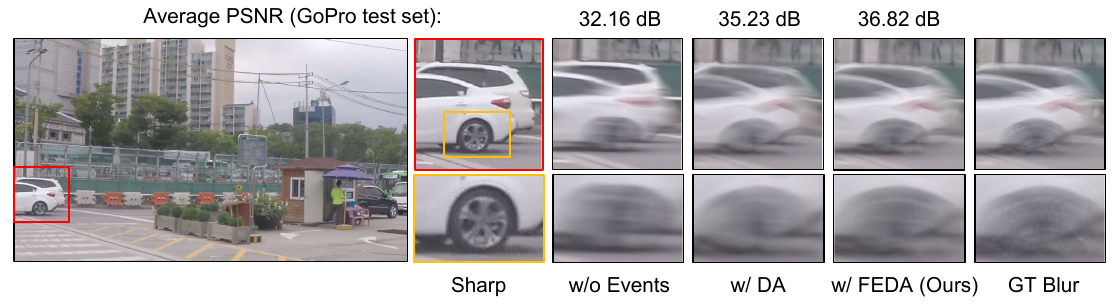}
    \vspace{-5mm}
    \caption{Visual comparison of blur synthesis under different conditioning. While the baseline using only flow maps (w/o Events) only captures macro-motions, and DA~\cite{maenet} struggles in motion-intensive areas, our FEDA successfully synthesizes blur patterns with high fidelity. The average PSNR scores on the GoPro test set (top) quantitatively show its superior blur synthesis quality.
    }
    \vspace{-2mm}
    \label{fig:ablation_eventblur}
\end{figure}

\noindent\textbf{Qualtitative Results.}
Fig.~\ref{fig:deblur_results} shows that EvBS consistently restores finer details across EVRB, REVD, and HighREV; additional backbones are provided in the supplementary material. 
Compared to the pre-trained baselines, our adapted models consistently reconstruct sharper images with superior detail preservation.
Fig.~\ref{fig:comparison_visual} further shows sharper results than recent domain-adaptive methods. 

Fig.~\ref{fig:motion_synthesis} visualizes our dual-pathway synthesis.
InCG uses the content flow $\mathcal{F}_C$, whereas ExCG transfers a separate motion flow $\mathcal{F}_M$ to produce $\tilde{\mathcal{F}}_C$. The transferred flow follows $\mathcal{F}_M$ rather than $\mathcal{F}_C$, confirming effective motion transfer and increased blur diversity.

\begin{table}[t]
\centering
\caption{Effectiveness of motion components. Combining both intrinsic and extrinsic motion yields the highest PSNR. Conversely, applying random motion offsets severely degrades performance, validating the necessity of our motion transfer algorithm.}
\vspace{-2mm}
\resizebox{0.7\linewidth}{!}{
\begin{tabular}{ccc|c}
\toprule
Intrinsic & Extrinsic & Random & PSNR (dB) \\
\midrule
& & & 25.41 \\
$\checkmark$ & & & 28.03 \\
& $\checkmark$ & & 28.14 \\
& & $\checkmark$ & 24.58 \\
$\checkmark$ & $\checkmark$ & & \textbf{28.64} \\
\bottomrule
\end{tabular}
} 
\label{tab:ablation_motion}
\end{table}

\vspace{-3mm}

\subsection{Ablation Studies}
\label{ssec:abs}
We conduct ablation studies on the REVD~\cite{FEVD} using MAENet~\cite{maenet}.

\noindent\textbf{Effects of $r$, $n$, and $[\tau_{low}, \tau_{high}]$.} Fig.~\ref{fig:ablation_hyperparams} analyzes the impact of key hyperparameters. First, the source selection ratio peaks at $r=0.15$; lower values yield insufficient training pairs, while higher ratios introduce label noise by selecting patches with residual blur. Second, $n=2$ extrinsic pairs yields the best performance. 
Increasing $n$ incorporates lower-ranked (low MCS) candidates with weaker alignment to the target motion pattern, thereby compromising adaptation efficacy.
Lastly, evaluating the orientation threshold reveals that the $[\tau_{low}, \tau_{high}] = [15^\circ, 30^\circ]$ range provides an effective balance between directional consistency and motion diversity. 
A narrower range may restrict diversity by filtering out valid motion sources, while a broader range introduces disparate motions that induce unrealistic transfer artifacts.




\noindent\textbf{Effectiveness of Motion Components.}
Table~\ref{tab:ablation_motion} summarizes the contributions of our Blur Conditioning Module (BCM). Compared to the pre-trained baseline (25.41 dB), fine-tuning with solely InCG (intrinsic) or ExCG (extrinsic) pairs improves PSNR to 28.03 dB and 28.14 dB, respectively. Combining both yields the best performance of 28.64 dB, demonstrating the synergistic effect of utilizing diverse motion priors in the target domain. 

To validate the ExCG design, we test a random offset baseline by randomly sampling $\mathbf{T}_k$ (Algorithm~\ref{alg:motion_transfer}). This drops the PSNR to 24.58 dB, even underperforming the baseline. This degradation highlights that our motion-aware transfer is crucial for synthesizing pairs aligned with the target domain distribution.

\noindent\textbf{Effectiveness of Event-guided Blurring.} We validate our FEDA representation against two conditioning baselines: using only flow maps (w/o Events, as in ID-Blau~\cite{IDBlau}), and with original deviation accumulation (w/ DA~\cite{maenet}) on GoPro test set. As shown in Fig.~\ref{fig:ablation_eventblur}, flow-only conditioning fails to accurately capture motion trajectories, whereas the DA captures motion trajectories but lacks capturing motion intensity. Conversely, our FEDA accurately models both trajectory and intensity, yielding the best blur-synthesizing quality (36.82 dB in PSNR across GoPro test set) that closely matches the ground truth. 

Lastly, when these synthesized pairs are used for domain-adaptive motion deburring, our FEDA-guided model achieves the highest performance (28.64 dB), compared to the adapted models trained with DA (28.53 dB) and flow-only (28.16 dB) conditionings. This confirms that FEDA provides richer motion cues, producing more effective training pairs for fine-tuning.


\vspace{-3mm}
\section{Conclusion}
In this paper, we propose EvBS, a novel event-guided diverse blur synthesis framework, to address the domain gap in motion deblurring. By leveraging events to decouple motion information from visual content, our framework synthesizes diverse training pairs through intrinsic and extrinsic motion synthesis strategies. These strategies produce richer training data that break the inherent constraints of naturally coupled motion and content. Comprehensive experimental results demonstrate that EvBS effectively enhances adaptation performance, significantly improving the robustness of state-of-the-art deblurring models on unseen domains.

\begin{acks}
This work was supported in part by Samsung Electronics Co., Ltd., System LSI Division, and by the InnoCORE program of the Ministry of Science and ICT (AI Meta-Scientist, No. N10260110).
\end{acks}
\vspace{-1.5mm}



\bibliographystyle{ACM-Reference-Format}
\bibliography{acmart}


\appendix
\clearpage

\twocolumn 
\begin{center}
    \vspace*{0.8cm} 
    {\Large \bfseries Supplementary Material}
    \vspace*{0.8cm} 
\end{center}


\newcommand{\appendixnumbering}{
    \renewcommand{\thefigure}{S\arabic{figure}}
    \renewcommand{\thetable}{S\arabic{table}}
    \renewcommand{\thealgorithm}{S\arabic{algorithm}}
}

\appendixnumbering
\setcounter{section}{0}

\setcounter{algorithm}{0}
\setcounter{page}{1}
\setcounter{table}{0}
\setcounter{figure}{0}

\begin{algorithm}
\caption{Flow-Enhanced Deviation Accumulation}
\begin{algorithmic}[1]
\item \textbf{Input:} Taking a pixel at $(x,y)$ as a specific example, $N$ Events stream $\{\mathbf{E}_0, \mathbf{E}_1, \cdots, \mathbf{E}_{N-1}\}$, Bidirectional flows $\{\mathbf{F}_{0\rightarrow1}, \mathbf{F}_{1\rightarrow2}, \cdots, \mathbf{F}_{N-1\rightarrow N} \}$, $\{\mathbf{F}_{0\rightarrow-1}, \mathbf{F}_{1\rightarrow0}, \cdots, \mathbf{F}_{N-1\rightarrow N-2} \}$. \\
\item \textbf{Output:} \leavevmode FEDA event representation $V$.
\STATE $R = 0$, $D = 0$, $C = 0$
\FOR{$i=0$ to $N-1$}
    \STATE Counting the number of events $C_i$ at $(x,y)$
    \FOR{$k=0$ to $C_i-1$}
        \STATE Event $E_i = (x, y, t_k, p_k)$
        \STATE Normalize timestamps: $dt_k = \frac{t_k - t_0}{t_{C_i-1} - t_0}$

        \STATE Compute $\mathbf{v}_0 = \frac{\mathbf{F}_{i \rightarrow i+1} - \mathbf{F}_{i \rightarrow i-1}}{2}$
        \STATE Compute $\mathbf{a} = \mathbf{F}_{i \rightarrow i+1} + \mathbf{F}_{i \rightarrow i-1}$
        \STATE Quadratic flow: $\mathbf{Q}_{i \rightarrow t_k} = \mathbf{v}_0 \cdot dt_k + 0.5 \cdot \mathbf{a} \cdot dt_k^2$
        \STATE Extract flow magnitude $|M_{t_k}|$ from $\mathbf{Q}_{i \rightarrow t_k}$
        

        \STATE $D = p_k \cdot |M_{t_k}|$
        \STATE $R = R + D$
    \ENDFOR
    \STATE $C = C + C_i$
\ENDFOR
\STATE Compute final FEDA representation: $V = R / C$
\end{algorithmic}
\label{alg:feda}
\end{algorithm}

\section{Flow-Enhanced Deviation Accumulation (FEDA)}
\label{sec:feda}
Deviation Accumulation (DA)~\cite{maenet} effectively encodes motion trajectory but accumulates events uniformly, disregarding motion intensity. To generate richer event-based conditioning signals, we propose FEDA, which enhances DA with quadratic flow~\cite{quadratic} magnitude information to better capture fine-grained motion dynamics.

Algorithm~\ref{alg:feda} details the FEDA process. Unlike standard deviation accumulation, FEDA weights each event by its corresponding motion magnitude. Specifically, for each event at $(x,y)$, we estimate the motion vector using quadratic flow, which explicitly models acceleration. The event polarity is then scaled by this computed motion magnitude ($|M_{t_k}|$) and accumulated. Finally, the sum is normalized by the total event count to yield a robust event representation $V$. This magnitude-weighted accumulation emphasizes motion-intensive regions more effectively, producing discriminative event representations.

We use linear magnitude weighting to preserve the natural proportionality between physical displacement and motion blur. Non-linear or learned weighting remains an alternative design choice, but may alter this relationship or add complexity.



\section{Motion Source Selector (MS Selector)}
\label{sec:msselector}
\begin{figure}[t]
    \centering
    \includegraphics[width=0.45\textwidth]{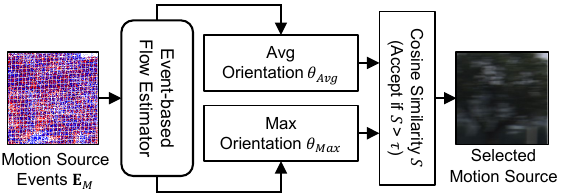}
    \caption{Illustration of the Motion Source Selector}
    \label{fig:MSselector}
\end{figure}

To ensure directional coherence in motion transfer, we introduce a Motion Source (MS) Selector (Fig.~\ref{fig:MSselector}) that evaluates the consistency of motion patterns present in the given motion source.
Specifically, we partition the event stream into two consecutive temporal segments and estimate the optical flow $\mathbf{F}=[u; v] \in \mathbb{R}^{2 \times H \times W}$ using event-based flow estimator~\cite{ERAFT}.

Using the horizontal and vertical components $u, v$, we compute the average orientation $\boldsymbol{\theta}_{\text{avg}} \in \mathbb{R}^2$ following the formulation in Eq.~\ref{eq:flow_avg}:
\begin{equation}
\bar{u} = \frac{1}{N}\sum u, \quad
\bar{v} = \frac{1}{N}\sum v, \quad
\boldsymbol{\theta}_{\text{avg}} = \frac{[\bar{u}; \bar{v}]}{\|[\bar{u}; \bar{v}]\|},
\label{eq:ms_avg}
\end{equation}
where $N$ is the number of pixels. Concurrently, we identify the dominant orientation $\boldsymbol{\theta}_{\text{max}} = [u; v]_{\mathbf{x}^*}\in \mathbb{R}^2$ at the location $\mathbf{x}^*$ with the maximum flow magnitude ($\|\mathbf{F}_{\mathbf{x}}\|_2$).
Finally, we calculate the cosine similarity $S = \boldsymbol{\theta}_{\text{avg}} \cdot \boldsymbol{\theta}_{\text{max}}$. We accept candidates only if $S > \tau$ (with $\tau=0.9$), ensuring that the selected source exhibits a coherent trajectory suitable for transfer.

\begin{figure}[t]
    \centering    
    \begin{subfigure}[b]{0.44\textwidth}
        \centering
        \includegraphics[width=1.0\textwidth]{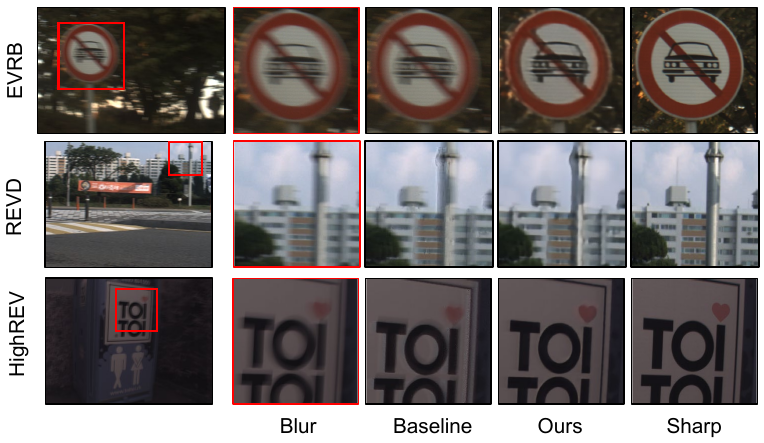} 
        \vspace{-1mm}
        \caption{EBFI}
    \end{subfigure}
    
    \begin{subfigure}[b]{0.44\textwidth}
        \centering
        \includegraphics[width=1.0\textwidth]{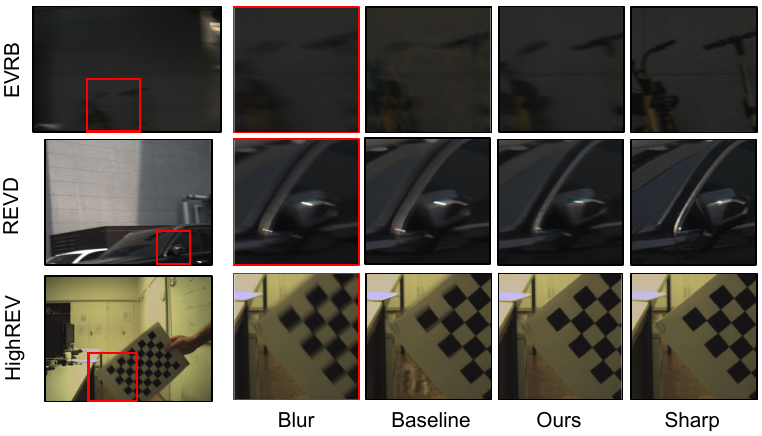} 
        \vspace{-1mm}
        \caption{NAFNet}
    \end{subfigure}

    \begin{subfigure}[b]{0.44\textwidth}
        \centering
        \includegraphics[width=1.0\textwidth]{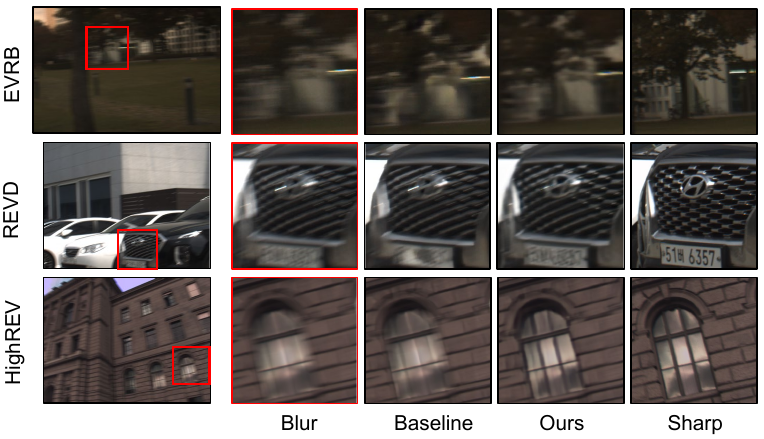} 
        \vspace{-1mm}
        \caption{MIMO-UNet}
    \end{subfigure}

    \vspace{-3mm}
    \caption{
    Additional qualitative results on EBFI, NAFNet, MIMO-UNet.
    We compare the blurry input (blur), pre-trained baseline (baseline), our adapted model (ours), and ground truth (sharp), with red boxes denoting zoomed-in regions.
    Our approach consistently restores finer details across both event-based (a) and frame-only (b, c) models.
    }
    \label{fig:morequal}
\end{figure}

\begin{figure}[t]
    \centering
    \includegraphics[width=0.5\textwidth]{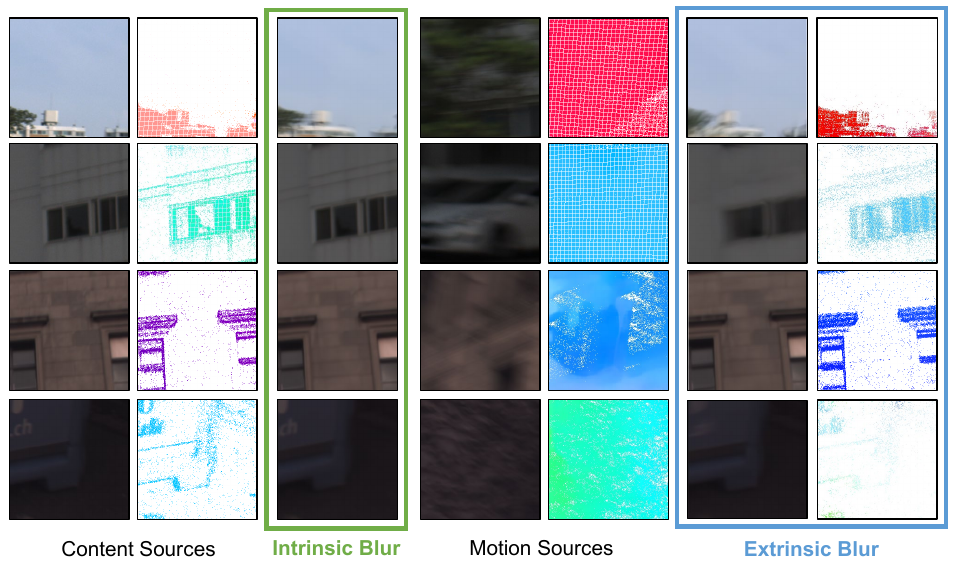}
    \vspace{-3mm}
    \caption{Additional qualitative examples of dual blur synthesis (intrinsic and extrinsic). Similar to Fig.~\ref{fig:motion_synthesis} in the main manuscript, each row displays a content source (left), followed by the intrinsic synthesized blur (green box) which utilizes inherent motion. The middle columns show the motion source, and lastly, the resulting extrinsic synthesized blur on the right (blue box).}
    \label{fig:morequal_motion}
\end{figure}


\section{More Qualitative Results}
\label{sec:more_qual}
We provide additional qualitative comparisons on EBFI~\cite{EBFI}, NAFNet~\cite{nafnet}, and MIMO-UNet~\cite{mimounet} in Fig.~\ref{fig:morequal}. While the baseline model struggles with residual blur and artifacts due to domain shift, our adapted version consistently produces sharper images with better-preserved fine details. This visual evidence further validates the robustness of our event-guided domain adaptation framework across diverse real-world datasets.

Furthermore, in Fig.~\ref{fig:morequal_motion}, we present additional visual examples of our blur synthesis process. These results illustrate our framework's ability to consistently generate diverse blurred patches across various scenes. As shown, our method successfully synthesizes both intrinsic blur, utilizing the content's inherent motion, and extrinsic blur, driven by motion patterns transferred from separate motion sources. This further confirms that our motion decoupling mechanism effectively expands the diversity of training pairs, which ultimately drives the enhanced adaptation performance.

\begin{table}[t]
\centering
\caption{Computational cost analysis (adapting on EVRB with MAENet, 
using RTX 2080 Ti).}
\label{tab:comp_cost}
\begin{tabular}{lc}
\toprule
\textbf{Component} & \textbf{Time (s)} \\
\midrule
Dual Source Extractor & 270.3 \\
\midrule
Blur Conditioning Module &  \\
\quad Intrinsic Condition Generator & 61.5 \\
\quad Extrinsic Condition Generator & 428.8 \\
\midrule
Blurring & 288.7 \\
Fine-tuning & 264.3 \\
\bottomrule
\end{tabular}
\end{table}

\section{Computational Cost}
\label{sec:comp}
Table~\ref{tab:comp_cost} details the computational cost of our framework on the EVRB dataset (894 images) using MAENet~\cite{maenet} with a single NVIDIA RTX 2080 Ti. The Dual Source Extractor takes 270.3s, while the Motion Conditioning Module requires 490.3s in total. Notably, the Extrinsic Condition Generator (ExCG) dominates the cost (428.8s) due to the computational complexity of motion pattern matching and transfer. The blurring and fine-tuning stages take 288.7s and 264.3s, respectively. 
Finally, it is worth noting that our adaptation process is conducted offline; thus, it does not affect the run-time efficiency of the deblurring model during inference.



\section{Limitations and Future Work}
We highlight three practical limitations of our framework. First, the source selection ratio $r$ should be chosen conservatively, as a large $r$ may include residual blur and introduce noisy supervision; empirically, $r=10$--$15\%$ provides a good trade-off (Fig.~\ref{fig:8a}). Second, ExCG incurs additional computational cost (428.8\,s vs. 61.5\,s for InCG on EVRB) for sometimes modest gains (e.g., $\sim$0.5\,dB on REVD); thus, InCG-only offers a faster alternative when adaptation time is constrained. Finally, EvBS relies on well-aligned event--image pairs, and performance may degrade under sensor misalignment. Improving robustness to such misalignment is an important direction for future work.

\end{document}